# FRASIMED: a Clinical French Annotated Resource Produced through Crosslingual BERT-Based Annotation Projection

Jamil Zaghir[1,2], Mina Bjelogrlic[1,2], Jean-Philippe Goldman[1,2], Soukaïna Aananou[1,2], Christophe Gaudet-Blavignac[1,2], Christian Lovis[1,2]

[1] Division of Medical Information Sciences (SIMED), University Hospitals of Geneva, Geneva, Switzerland
[2] Department of Radiology and Medical Informatics, University of Geneva, Geneva, Switzerland
{jamil.zaghir, mina.bjelogrlic, jean-philippe.goldman, soukaina.aananou, christophe.gaudet-blavignac, christian.lovis}@unige.ch

**Abstract**
Natural language processing (NLP) applications such as named entity recognition (NER) for low-resource corpora do not benefit from recent advances in the development of large language models where there is still a need for larger annotated datasets. This research article introduces a methodology for generating translated versions of annotated datasets through crosslingual annotation projection. Leveraging a language agnostic BERT-based approach, it is an efficient solution to increase low-resource corpora with few human efforts and by only using already available open data resources. Quantitative and qualitative evaluations are often lacking when it comes to evaluating the quality and effectiveness of semi-automatic data generation strategies. The evaluation of our crosslingual annotation projection approach showed both effectiveness and high accuracy in the resulting dataset. As a practical application of this methodology, we present the creation of French Annotated Resource with Semantic Information for Medical Entities Detection (FRASIMED), an annotated corpus comprising 2'051 synthetic clinical cases in French. The corpus is now available for researchers and practitioners to develop and refine French natural language processing applications in the clinical field (https://zenodo.org/record/8355629), making it the largest open annotated corpus with linked medical concepts in French.

## 1. Introduction

The field of natural language processing (NLP) has witnessed significant advancements due to the availability of large, annotated corpora. These corpora serve as valuable resources for training and evaluating machine learning models, enabling the development of robust and accurate NLP applications such as Named Entity Recognition (NER). However, creating high-quality annotations for large datasets is a time-consuming and resource-intensive task. Manual annotation requires domain experts, which can be particularly challenging when dealing with specialized domains.

To address this challenge, researchers have explored automated and semi-automated methods for generating annotations, aiming to reduce the human effort involved in the annotation process. One promising approach is annotation projection, which involves transferring annotations from one language to another. By leveraging the existing annotations in a different language, it provides an efficient way of creating annotated datasets in languages where such annotations are limited or nonexistent.

## 2. Related works

### 2.1 Crosslingual annotation projection

Several studies have explored the use of crosslingual annotation projection techniques to save annotation time, ensure comparability across languages, and create valuable multilingual resources.

Bentivogli and Pianta (2005) proposed an approach to create linguistically annotated resources based on aligned parallel corpora. By transferring annotations from a source language to a target language using a bilingual knowledge-intensive word aligner, they were able to reduce human effort and exploit existing annotated resources to create new ones. This is exemplified in the creation of the MultiSemCor corpus, an English-Italian semantically annotated parallel corpus based on translations of the English SemCor corpus.

Previous contributions used crosslingual annotation projection, either to create a new multilingual resource or to perform NER. The methods are based on the use of a Machine Translation (MT) system, followed by variants of string matching: Ehrmann et al. (2011), Jain et al. (2019) and Miranda-Escalada et al. (2022) respectively used consonant signature matching (i.e., a variant of string matching but with vowels removal), phonetic similarity, and lemmatized string matching. Alternatively, Abacha et al. (2013) projected annotations using MOSES (Koehn et al. 2007), a word alignment tool based on a statistical model. The evaluation of the projected annotations is indirectly done quantitatively through the evaluation of the final NER task.

Furthermore, annotation projections have been applied to other tasks such as Semantic Role Labeling and Part-of-Speech Tagging (Padó and Lapata 2009; Van der Plas et al. 2011, 2014; Huck et al. 2019; Aminian et al. 2019; Daza and Frank 2020). Knowledge-based annotation projection have also been proposed: Padó and Lapata (2009) used FrameNet (Baker 2017) to create an English-German parallel corpus with high-quality annotations, with minimal human effort. Van der Plas et al. (2011) generated semantic role annotations for the French language using PropBank (Kingsbury and Palmer 2002), an English corpus annotated with semantic propositions.

To the best of the authors' knowledge, no previous work has utilized a Transformer-based annotation projection to produce a new annotated dataset. While these previous studies have demonstrated the feasibility of crosslingual annotation projection techniques, the creation of French annotated medical corpora through annotation projection remains largely unexplored.

## 2.2 Available French medical NER data

The French research community has previously collected and distributed clinical and biomedical datasets for NER.

One such dataset is CAS corpus (Grabar et al. 2018), a collection incorporating clinical cases originating from French scientific literature. Annotated with demographic attributes (age, gender) and clinical features (origin of visit, outcome), the corpus encompasses 717 clinical documents, comprising approximately 231'662 tokens.

A subset of CAS, known as DEFT, emerged within the 2020 DEFT challenge (Cardon et al. 2020). Comprising 167 cases, DEFT integrates 13'493 annotations with 13 different clinical entity types.

The QUAERO dataset features a French biomedical NER Gold Standard (Névéol et al. 2014). Drawing from PubMed article titles and European MEdicines Agency (EMEA) documents, it boasts manual annotations encompassing ten clinical entity categories, all linked to the Unified Medical Language System (UMLS) Metathesaurus. Spanning 1'688 files, this corpus accommodates 26'409 UMLS-linked annotations.

The European Clinical Case Corpus (E3C) (Magnini et al. 2021) stands out as a multilingual resource encompassing English, French, Italian, Spanish, and Basque. It offers annotated clinical narratives, aligning clinical entities (disorders) with the UMLS terminology and incorporating temporal information. Comprising 81 French documents with 29'256 tokens, E3C comprises 1'327 annotated entities, with 1'085 linked to UMLS concepts.

Finally, Miranda-Escalada et al. (2022) presented DISTEMIST, a Spanish corpus comprising 1'000 synthetic clinical cases annotated with diseases linked to SNOMED-CT (SNOMED International 1999). They produced a multilingual dataset by employing a lemmatized string-matching based annotation projection method across six languages, including French. However, the evaluation of the projections' quality is limited.

While these datasets collectively provide a rich foundation for advancing clinical natural language processing, the number of very large French clinical NER datasets available to the public is limited.

## 3. Contributions of this study

This paper builds upon this existing line of research and presents a novel methodology for generating translated versions of annotated datasets using a language-independent BERT-based approach (Devlin et al. 2018). The proposed annotation projection methodology offers several advantages: (i) it does not require any ontology or lexical resource, (ii) it is applicable to multiple languages, including low-resource ones, and (iii) publicly available fine-tuned BERT models are used for crosslingual annotation projection.

As a practical application of our methodology, the focus is on creating a French annotated corpus, named FRASIMED, through MT and annotation projection. FRASIMED originates from two Spanish annotated corpora consisting of synthetic cases: the first corpus, CANTEMIST, includes oncological cases (Miranda-Escalada et al. 2020), while the second corpus, DISTEMIST, contains annotated diseases (Miranda-Escalada et al. 2022). An example of annotation from the original CANTEMIST corpus is presented in Figure 1.

The contributions of this work are summarized as follows:
- A novel pipeline for generating translated versions of annotated datasets using a language-independent, crosslingual annotation projection approach.
- A novel corpus, FRASIMED, composed of synthetic clinical cases for French NER and Entity Linking freely available at https://zenodo.org/record/8355629, making it the largest open annotated corpus for oncological concepts in French.
- A set of extensive experiments to evaluate the annotation projection quantitatively and qualitatively.

## 4. Methodology

In this section, we introduce our methodology which consists of the following steps, illustrated in Figure 2:
- First, we use an existing document-level MT system to get the translated version of the textual content, as detailed in Section 4.2.
- Next, an automatic annotation projection pipeline is run to generate the annotation files for the translated text. Also, we produce a TSV file gathering the projected annotations, explained in Section 4.3.
- Then, using this report we semi-automatically detect badly projected annotations by going through this report - presented in Section 4.4.
- Finally, we manually correct these badly projected annotations, as discussed in Section 4.5.

## 4.1 Datasets

We produce a French version of a dataset made available through CANTEMIST (CANcer TExt Mining Shared Task), a shared task focusing on NER and Entity Linking of Spanish clinical texts in oncology (Miranda-Escalada et al. 2020). Through a publicly available corpus of 1'301 annotated synthetic cases, the task aims to automatically detect mentions of tumor morphology terms, as well as linking them to a medical terminology. As illustrated in Figure 1, each entity in this dataset is associated with a morphology code from the International Classification of Diseases for Oncology (ICD-O-3.1) (WHO 2013).

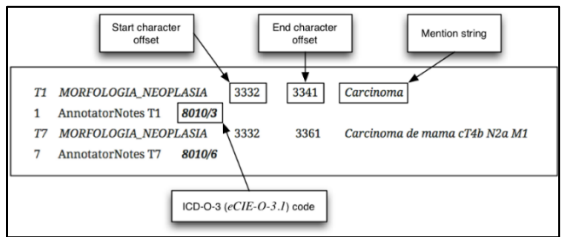

Figure 1: An example of CANTEMIST annotation

Moreover, in addition to the CANTEMIST dataset, we also include a subset from a dataset named DISTEMIST (DISease TExt Mining Shared Task). This subset consists of 750 cases, and it is annotated with diseases (Miranda-Escalada et al. 2022), with linkages to the July 31, 2021 release of SNOMED-CT. The combination of these two datasets, CANTEMIST and DISTEMIST, allows us to create a more comprehensive French annotated corpus for medical NLP tasks. Also, by including the DISTEMIST dataset, we can conduct a comparative analysis between our methodology and the annotation projection method employed by the original authors.

### 4.2 Document-level MT

Based on previous works (Läubli et al. 2018; Sun et al. 2022), it has been established that document-level translation is a more effective approach in comparison to sentence-level translation. This is primarily attributed to its capacity to capture the contextual nuances and overall coherence of a document. Nevertheless, the adoption of this approach comes at the expense of losing the information pertaining to sentence alignment. Document-level translation takes into consideration the broader context encompassing themes, subject matter, and intended meaning of the entire document. The precise reason for the superiority of document-level neural MTs compared to those involving isolated sentences remains unclear: Kim et al. (2019) discovered that most improvements are not readily interpretable.

In our study, we use DeepL Pro to translate Spanish documents into French. This step has been applied to CANTEMIST. However, for DISTEMIST, we opt to directly use the semi-automatically translated French texts provided by the original authors.

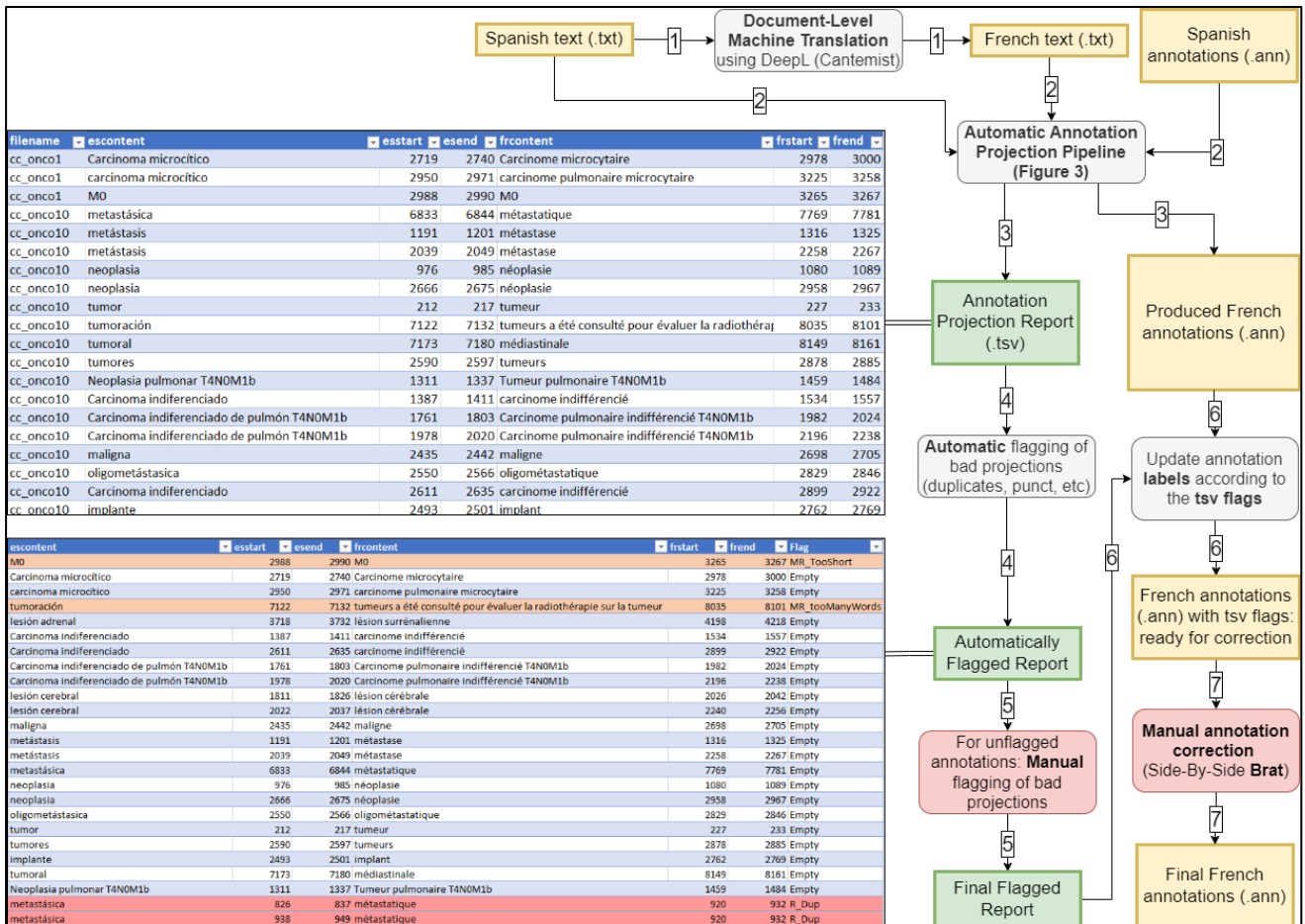

Figure 2: The comprehensive workflow describing our semi-automatic methodology (green and yellow squares depicts the TSV file, and the corpus respectively, red and grey rounded boxes are automatic and manual tasks respectively).

## 4.3 Annotation projection pipeline

After having obtained the translated version of the unannotated text, we launch a pipeline that aims to project annotations from one language to another. As illustrated in Figure 3, the process begins with a textual document, its translated version, and its annotation file.

**Sentence-level pipeline**: Firstly, the Spanish document and its corresponding French translation are split into individual sentences. To represent each sentence effectively, a Language Agnostic BERT Sentence Embedding (LABSE) (Feng et al. 2022) is employed[1]. This BERT model is trained on both monolingual and bilingual corpora, employing Masked Language Modeling (MLM) and Translation Language Modeling (TLM) as pre-training objectives. MLM requires the model to predict the [MASK] token in the context of surrounding tokens, while TLM extends this objective to a bilingual context by inputting a concatenated translation pair. With support for over 109 languages, this model has achieved state-of-the-art performance in bi-text retrieval. Notably, LABSE surpasses multilingual BERT in generating superior crosslingual sentence embeddings, primarily due to TLM's focus on sentence-level objective. By comparing the sentence representations using similarity measures, sentence alignment is established, determining which sentences correspond between the two languages (Liu and Zhu 2023). The code aligning bilingual sentences using LABSE is publicly available on GitHub[2].

**Word-level pipeline**: Once the sentence alignment is established, the focus shifts to word-level alignment. Each pair of aligned sentences is further processed by splitting the words. The word alignment is then computed by encoding the words using multilingual BERT and measuring the cosine similarity between them: each word in the source sentence is aligned with the closest word in the target one. To account for the one-to-many relationships between words, the word alignment is applied bidirectionally (Figure 4). This means that the alignment is performed both from Spanish to French and from French to Spanish, ensuring comprehensive coverage of the annotations. The code aligning words from a bilingual sentence pair using multilingual BERT is also publicly available on GitHub[3].

**Annotation projection**: Finally, leveraging the results of the word alignment, the French annotation file is constructed from the original Spanish annotation file. In cases where the resulting annotation projection exhibits discontinuity, a gluttonous strategy is employed: this approach involves annotating all the words intervening between the fragmented annotations to form a cohesive and continuous annotation. This approach aims to prevent fragmented annotations while favoring the recall.

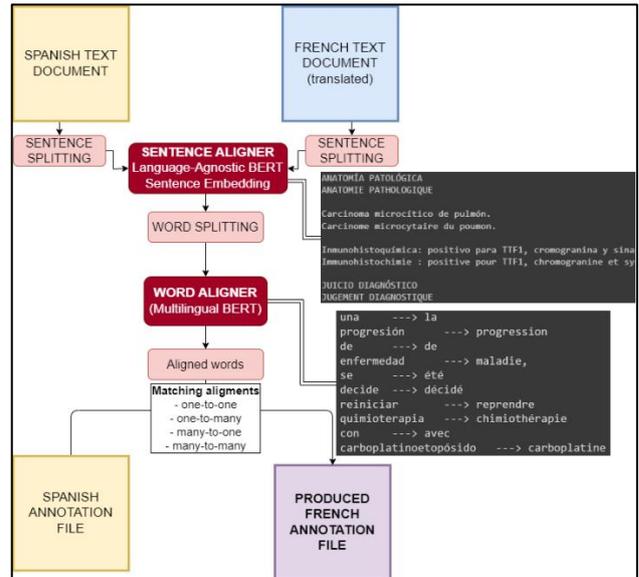

Figure 3: Automatic annotation projection pipeline

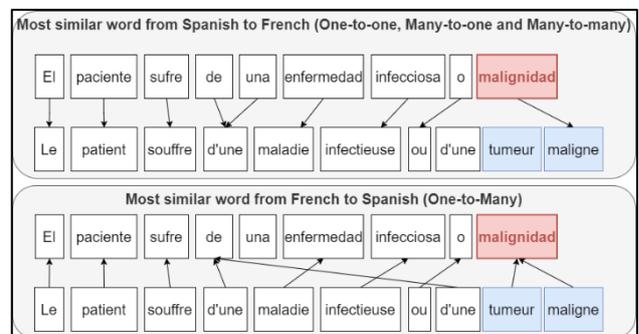

Figure 4: Bidirectional word alignment using cosine similarity. Red box is the Spanish annotation, blue boxes are the expected annotation projection.

## 4.4 Identification of badly projected annotations

Upon the successful execution of the automated annotation projection pipeline, a report is generated in TSV format, enumerating the crosslingual annotation projections that were conducted. The purpose of this report is to present a comprehensive examination of potential errors that may arise during the annotation projection process and facilitate their identification.

The initial phase involves discerning distinct error patterns that may manifest. These patterns include, among others:
- instances of duplicate annotations.
- presence of punctuation in projected annotations while it was originally absent.
- annotations of excessively short length.
- undesirable words such as coordinating conjunctions and determiners appearing at the beginning or end of the annotations.
- absence of alphanumeric characters in the projected annotations.

---

[1] https://github.com/bojone/labse
[2] https://github.com/bfsujason/bertalign
[3] https://github.com/andreabac3/Word_Alignment_BERT

- a relatively higher word count in the projected annotations compared to the original ones.

The subsequent phase involves the identification of annotations that appear only once throughout the entire corpus, rendering them initially questionable. The aim is to manually review these suspicious annotations to either validate or disprove the uncertainty. For both datasets, which encompassed a total of 24,097 projected annotations, this methodology proved efficient.

Lastly, the generated annotation files' tags are adjusted based on the outcomes from the TSV file, as a preparation step for the manual correction of these annotations.

### 4.5 Manual annotation correction

Using the tags obtained from the error identification of projections detailed in the preceding section, it becomes feasible to perform a manual correction of the corpus in an efficient manner.

We establish an annotator guideline encompassing the tag meanings and instructions to handle complex annotation projections. For instance, it addresses scenarios caused by the translation phenomena, such as revealing the subject in French that was initially hidden in Spanish.

The manual correction process is carried out employing Brat (Stenetorp et al. 2012), a web-based tool specifically designed for textual annotation. Moreover, Brat enables the side-by-side display (Figure 5), considerably expediting the correction of crosslingual annotation projections. The correction of CANTEMIST and DISTEMIST datasets required 28 and 20 hours of annotations, respectively.

## 5. Experiments and results

We provide a comprehensive overview of our experimental approach by encapsulating the entire paper's workflow in the Figure 6. This figure illustrates the strategic allocation of experiments across various stages of the workflow, allowing for a concise representation of our research methodology and the specific components targeted by each experiment.

### 5.1 Quality of the automatic annotation projection pipeline

To evaluate the performance of our automatic annotation projection pipeline, we conduct a comparative analysis between the annotation files generated by the pipeline before manual correction and the manually corrected versions (i.e., the gold standard). The evaluation metrics used in this analysis include:
- CORRECT: Entities identified in both sets, with matching annotation indices.
- PARTIAL: Cases where entities are partially identified (non-matching annotation indices).
- MISSING: Instances where the system fails to identify an expected entity, present in the gold standard.
- SPURIOUS: Incorrect identification of entities, i.e., when the system identifies a word or phrase as an entity erroneously, absent in the gold standard.

The Table 1 shows that CORRECT annotations are the most prevalent, followed by PARTIAL annotations. Given that several PARTIAL annotations involve minor corrections such as punctuations or single-character differences, we also introduce a relaxed version of Precision, Recall, and F1-Score to consider these annotations as valid along with the CORRECT ones. The presence of SPURIOUS and MISSING annotations underscores the importance of the manual correction procedure.

| | CANTEMIST | DISTEMIST |
|---|---|---|
| **CORRECT** | 13'404 | 5'991 |
| **PARTIAL** | 2'701 | 2'376 |
| **MISSING** | 770 | 333 |
| **SPURIOUS** | 424 | 228 |
| **PRECISION (Relaxed)** [CORRECT+PARTIAL] / [CORRECT+PARTIAL+ SPURIOUS] | 97.4% | 97.3% |
| **RECALL (Relaxed)** [CORRECT+PARTIAL] / [CORRECT+PARTIAL+ MISSING] | 95.4% | 96.1% |
| **F1-SCORE (Relaxed)** | 96.4% | 96.8% |
| **PRECISION (Strict)** [CORRECT] / [CORRECT +PARTIAL+ SPURIOUS] | 81.1% | 69.6% |
| **RECALL (Strict)** [CORRECT] / [CORRECT +PARTIAL+MISSING] | 79.4% | 68.8% |
| **F1-SCORE (Strict)** | 80.2% | 69.2% |

Table 1: Evaluation results of BERT projection against the manual correction as the gold standard

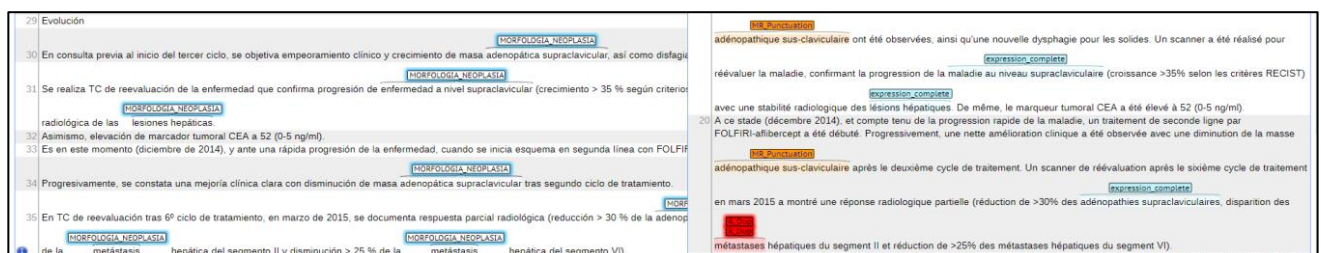

Figure 5: Brat in side-by-side display mode. The Spanish annotation file is displayed on the left, and the French version on the right. The red / orange label means the annotation is false / suspicious, respectively.

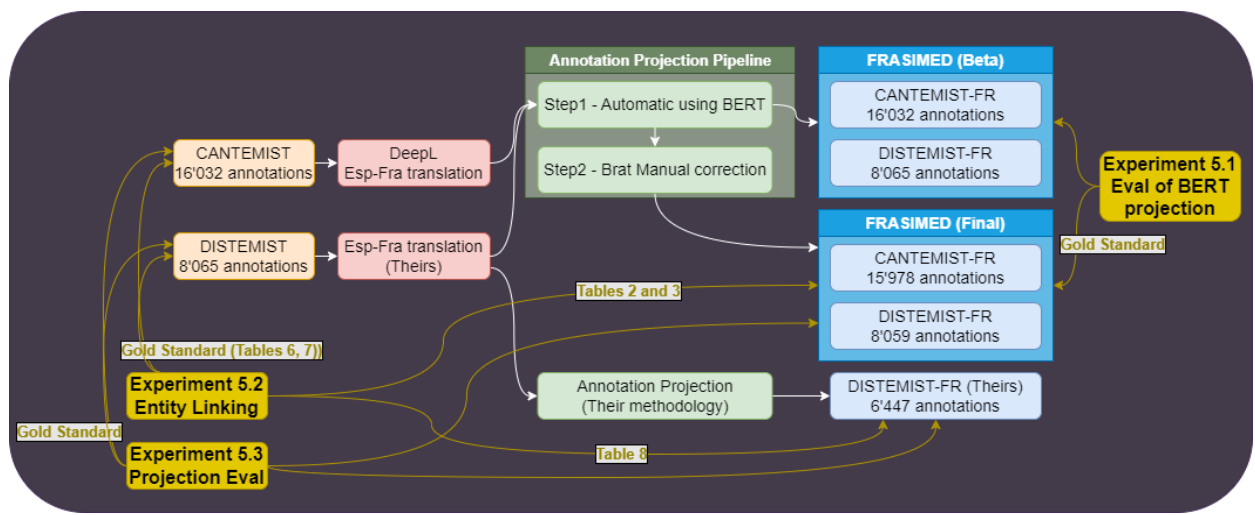

Figure 6: Summary of experiments (golden arrows and nodes are experiments-related, while white arrows refer to the pipeline's process flow. Orange and blue nodes are Spanish and French annotated corpora, resp.)

## 5.2 Quality of entity linking

In this experiment, we examine the quality of entities linked to medical terminologies in various datasets, including:

- Our French version of CANTEMIST (Table 2)
- Our French version of DISTEMIST (Table 3)
- The original Spanish CANTEMIST (Table 6 in the Appendix)
- The original Spanish DISTEMIST (Table 7 in the Appendix)
- The French version of DISTEMIST created by the original authors through their annotation projection method (Table 8 in the Appendix)

| ICD-O | ICD-O Concept description | Count | Most occurrent |
|---|---|---|---|
| 8000/6 | Neoplasm, metastatic | 5721 | métastases |
| 8000/1 | Neoplasm, uncertain whether benign or malignant | 2805 | tumeur |
| 8000/3 | Neoplasm, malignant | 966 | cancer |
| 8140/3 | Adenocarcinoma, NOS (not otherwise specified | 487 | adénocarcinome |
| 8010/3 | Carcinoma, NOS | 223 | carcinome |
| 8500/3 | Infiltrating duct carcinoma, NOS | 214 | carcinome canalaire infiltrant |
| 8140/6 | Adenocarcinoma, metastatic, NOS | 212 | métastase d'adénocarcinome |
| 8720/3 | Malignant melanoma, NOS | 166 | mélanome |
| 8010/9 | Carcinomatosis | 135 | carcinomatose |
| 8001/1 | Tumor cells, uncertain whether benign or malignant | 131 | cellules tumorales |

Table 2: Top 10 concepts in the French CANTEMIST dataset, with the most occurrent annotation found.

For the FRASIMED corpora (Tables 2 and 3), we observe proper alignment and consistency between concepts and the most frequently occurred annotations. These tables closely resemble their Spanish counterparts (Tables 6 and 7), with identical top 10 concepts.

| SNOMED | SNOMED Concept description | Count | Most occurrent |
|---|---|---|---|
| 417163006 | Traumatic or non-traumatic injury (disorder) | 146 | lésion |
| 38341003 | Hypertensive disorder, systemic arterial (disorder) | 94 | hypertension artérielle |
| 55342001 | Neoplastic disease (disorder) | 78 | tumeur |
| 157670007 | Injury NOS (disorder) | 51 | lésions |
| 30746006 | Lymphadenopathy (disorder) | 46 | adénopathie |
| 77176002 | Smoker (finding) | 41 | fumeur |
| 441457006 | Cyst (disorder) | 34 | kyste |
| 40733004 | Infectious disease (disorder) | 33 | infection |
| 128462008 | Metastatic malignant neoplasm (disorder) | 30 | métastase |
| 44054006 | Diabetes mellitus type 2 (disorder) | 26 | diabète sucré de type 2 |

Table 3: Top 10 concepts in our French DISTEMIST dataset, with the most occurrent annotation found.

Furthermore, we compare our French DISTEMIST dataset with the one created by the original authors through a lemmatization-based projection of annotations. Table 8 in the Appendix highlights that the most frequent concept has a significantly higher occurrence count, and its most occurred annotation is inconsistent. This is likely caused by the translation and lemmatization of the word "HTA", leading to "et", a French coordinating conjunction.

Additionally, the top 10 concepts do not align with the Spanish dataset, suggesting that the authors likely detected additional SNOMED concepts and missed those already present in the original Spanish corpus.

## 5.3 Comparison and evaluation of the full projection pipeline

### 5.3.1 Quantitative Evaluation

In this comparative study, we evaluate the quantitative aspects of entities and codes linked to medical terminologies across five datasets (Table 4).

Our analysis focuses on the original datasets and their projected counterparts. Both versions of CANTEMIST demonstrates that all entities were appropriately linked to an ICD-O code, signifying the preservation of entity-code associations through our annotation projection methodology. The slight difference in entity and code quantities between the original and projected dataset was expected due to manual corrections.

We also examine the French versions of the DISTEMIST dataset, namely DISTEMIST-FR (Miranda-Escalada et al. 2022), and DISTEMIST-FR (FRASIMED), in comparison to the original Spanish DISTEMIST-ES dataset. Like the CANTEMIST-FR dataset, our annotation projection methodology maintains consistency by retaining entities and codes from the original DISTEMIST-ES dataset and DISTEMIST-FR (FRASIMED). In contrast, the original DISTEMIST authors' alternative approach, using string-matching with translation and lemmatization, overlooked some entities. However, their method identified novel links to SNOMED terminologies, indicating a trade-off between entity coverage and new code associations.

| Dataset | # Entities | # Codes |
|---|---|---|
| CANTEMIST-ES (Miranda-Escalada et al. 2020) | 16'032 | 16'032 |
| **CANTEMIST-FR (FRASIMED)** | 15'978 | 15'978 |
| DISTEMIST-ES (Miranda-Escalada et al. 2022) | 8'065 | 5'136 |
| DISTEMIST-FR (Miranda-Escalada et al. 2022) | 6'447 | 6'445 |
| **DISTEMIST-FR (FRASIMED)** | 8'059 | 5'132 |

Table 4: Quantitative comparison between datasets

### 5.3.2 Qualitative Evaluation

The Table 5 presents the results of the qualitative evaluation conducted on 640 annotation projection pairs extracted from both our annotation projection pipeline and the one used by the original DISTEMIST authors. The choice of this sample size was deliberate, ensuring a sufficient level of statistical robustness with a 99% interval confidence and a margin of error of 5%.

During the qualitative analysis, a trained medical student was engaged to meticulously assess and analyze instances of poor annotation projections within these pairs. The evaluation specifically aimed to identify and categorize various types of problematic projections, shedding light on both the strengths and weaknesses of the two approaches.

The Table 5 reports the findings in three distinct categories:

**Bad projections:** These instances arise when the annotation projection results in semantically inaccurate associations between entities, often due to missing words in the projected annotations. The relatively low frequency of such occurrences attests to the effectiveness of both annotation projection algorithms in establishing accurate associations between entities. The primary cause of these omissions can be attributed to the limitations of MT in fully translating certain expressions.

**Bad translations caused by abbreviations** and **other bad translations:** In these cases, the annotation projection successfully establishes correct associations between entities. However, the translations are subpar due to the MT system's limitations in handling abbreviations or other factors unrelated to abbreviations, such as the use of unusual words in the clinical field or mistranslation of technical terms.

Comparing the annotation projection quality between the original authors of DISTEMIST and our approach, the original authors demonstrate superior annotation projection quality. Their string-matching based method prioritizes precision over recall, accurately detecting entities but potentially missing some in the process. Indeed, their approach results in fewer bad projections related to translation errors.

Conversely, in the case of FRASIMED, the alignment-based approach aims to match each concept, even if translations are suboptimal, leading to a higher number of mistranslated annotations. Instances of erroneous annotation projections, independent to mistranslations, constitute less than 2% of the total annotations.

| Dataset created with projections | Bad projections | Bad translations caused by abbreviations | Other bad translations |
|---|---|---|---|
| CANTEMIST (FRASIMED) | 11 (1.72%) | 22 (3.44%) | 31 (4.84%) |
| DISTEMIST (Miranda-Escalada et al. 2022) | **2 (0.31%)** | **2 (0.31%)** | **9 (1.41%)** |
| DISTEMIST (FRASIMED) | 7 (1.09%) | 7 (1.09%) | 35 (5.5%) |

Table 5: Qualitative evaluation of the projections, on 640 projection pairs for each dataset

## 6. FRASIMED Enrichment and Release

**Data augmentation through ICD-O-3 to SNOMED-CT mappings.** As SNOMED-CT is a versatile terminology with the ability to establish mappings with diverse international terminologies and classifications, we improve the French CANTEMIST dataset by adding the SNOMED-CT links. This is achieved through the mapping operation from ICD-O-3.1 to SNOMED-CT utilizing the UMLS Metathesaurus (Unified Medical Language System) as a pivotal terminological framework. This additional mapping allowed the addition of 15'457 new SNOMED codes (Table 9 in the Appendix). The whole dataset contains 24'037 annotations, with links to 15'978 ICD-O and 20'589 SNOMED codes.

**Dissemination of FRASIMED.** In a spirit of fostering research endeavors in French NLP, specifically in the domains of Clinical Medical Entity Recognition and Entity Linking to terminologies such as SNOMED-CT and ICD-O-3, we have made FRASIMED, encompassing the CANTEMIST and DISTEMIST datasets, accessible to the public. To the best of the authors' knowledge, this is the first large French dataset annotated with ICD-O and SNOMED concepts that has been made publicly available. This initiative aims to catalyze advancements in the French NLP field and encourage researchers to make novel contributions.

## 7. Conclusions

In conclusion, this study has introduced a robust methodology centered around crosslingual annotation projection, facilitating the generation of translated versions of annotated datasets. Employing a language-agnostic BERT-based approach, this method not only offers an efficient solution for projecting annotations across diverse languages, including those with limited resources, but also presents a practical case for expediting the creation of annotated datasets with preserved semantic and domain-specific information.

Through the practical application of this methodology, exemplified by the development of FRASIMED, we have achieved notable efficiencies in terms of time and resources. The technique's ability to transfer annotations from a source language to a target language obviates the need for manual annotation from scratch, thereby highlighting its potential to create a large resource with minimal effort, including for low-resource languages.

Our extensive experimentation has thoroughly assessed the quality and efficacy of the crosslingual annotation projection approach. While mistranslations are hurdles, it is crucial to note that the methodology of our semi-automatic annotation projection pipeline remains independent from the MT system. Its applicability extends to any bilingual corpus. Hence, we posit that the efficacy of our approach would have been further pronounced if the bilingual sentence pairs were not produced through MT.

With the release of the French annotated resource, brought into existence via this methodology, this work does not only contribute to advancing crosslingual annotation projection as a valuable tool in NLP but also fosters advancements in linguistic research and applications in the clinical French NLP. With this resource as a starting point, we encourage researchers to make future contributions, particularly to the automatic detection of SNOMED-CT concepts in French free-text.

## 8. Data Availability

FRASIMED, composed of clinical synthetic cases for French NER and Entity Linking, is freely available, under the Creative Commons 4.0 License, in Zenodo: https://doi.org/10.5281/zenodo.8355629.

## 9. Acknowledgements

This work was supported as a part of NCCR Evolving Language, a National Centre of Competence in Research, funded by the Swiss National Science Foundation (grant number **#51NF40_180888**). We would like to thank Vincent Guyot for his medical expertise required for the evaluation.

## 10. Bibliographical References

# Appendix: Tables for Experiment 5.2

| ICD-O | ICD-O Concept description | Count | Most occurent |
|---|---|---|---|
| 8000/6 | Neoplasm, metastatic | 5736 | metástasis |
| 8000/1 | Neoplasm, uncertain whether benign or malignant | 2811 | tumor |
| 8000/3 | Neoplasm, malignant | 974 | malignidad |
| 8140/3 | Adenocarcinoma, NOS (not otherwise specified) | 488 | adenocarcinoma |
| 8010/3 | Carcinoma, NOS | 223 | carcinoma |
| 8500/3 | Infiltrating duct carcinoma, NOS | 217 | carcinoma ductal infiltrante |
| 8140/6 | Adenocarcinoma, metastatic, NOS | 212 | metástasis de adenocarcinoma |
| 8720/3 | Malignant melanoma, NOS | 166 | melanoma |
| 8010/9 | Carcinomatosis | 136 | carcinomatosis |
| 8001/1 | Tumor cells, uncertain whether benign or malignant | 132 | células tumorales |

Table 6: Top 10 concepts in the Spanish CANTEMIST dataset, with the most occurrent annotation in the corpus

| SNOMED | SNOMED Concept description | Count | Most occurrent |
|---|---|---|---|
| 417163006 | Traumatic or non-traumatic injury (disorder) | 147 | lesión |
| 38341003 | Hypertensive disorder, systemic arterial (disorder) | 95 | hipertensión arterial |
| 55342001 | Neoplastic disease (disorder) | 78 | tumor |
| 157670007 | Injury NOS (disorder) | 51 | lesiones |
| 30746006 | Lymphadenopathy (disorder) | 46 | adenopatías |
| 77176002 | Smoker (finding) | 41 | fumador |
| 441457006 | Cyst (disorder) | 34 | quiste |
| 40733004 | Infectious disease (disorder) | 33 | infección |
| 128462008 | Metastatic malignant neoplasm (disorder) | 30 | metástasis |
| 44054006 | Diabetes mellitus type 2 (disorder) | 26 | diabetes mellitus tipo 2 |

Table 7: Top 10 concepts in the Spanish DISTEMIST dataset, with the most occurrent annotation in the corpus

| SNOMED | SNOMED Concept description | Count | Most occurrent |
|---|---|---|---|
| 38341003 | Hypertensive disorder, systemic arterial (disorder) | 385 | et |
| 417163006 | Traumatic or non-traumatic injury (disorder) | 297 | lésion |
| 55342001 | Neoplastic disease (disorder) | 159 | tumeur |
| 157670007 | Injury NOS (disorder) | 107 | lésions |
| 128462008 | Metastatic malignant neoplasm (disorder) | 78 | métastases |
| 441457006 | Cyst (disorder) | 70 | kyste |
| 30746006 | Lymphadenopathy (disorder) | 60 | adénopathies |
| 385494008 | Hematoma (disorder) | 52 | hématome |
| 40733004 | Infectious disease (disorder) | 46 | infection |
| 77176002 | Smoker (finding) | 44 | fumeur |

Table 8: Top 10 concepts in the French DISTEMIST dataset released by the original authors.

| SNOMED | SNOMED Concept description | Count | Most occurrent |
|---|---|---|---|
| 14799000 | Neoplasm, metastatic (morphologic abnormality) | 5727 | métastases |
| 86251006 | Neoplasm of uncertain behavior (morphologic abnormality) | 2856 | tumeur |
| 1240414004 | Malignant neoplasm (morphologic abnormality) | 1004 | cancer |
| 1187332001 | Adenocarcinoma (morphologic abnormality) | 658 | adénocarcinome |
| 58477004 | Infiltrating ductular carcinoma (morphologic abnormality) | 332 | carcinome canalaire infiltrant |
| 1187425009 | Carcinoma (morphologic abnormality) | 307 | carcinome |
| 4590003 | Adenocarcinoma, metastatic (morphologic abnormality) | 228 | métastase d'adénocarcinome |
| 1162635006 | Malignant melanoma (morphologic abnormality) | 168 | mélanome |
| 1162767002 | Squamous cell carcinoma (morphologic abnormality) | 168 | carcinome épidermoïde |
| 307593001 | Carcinomatosis (disorder) | 135 | carcinomatose |

Table 9: Top 10 concepts in the French CANTEMIST dataset regarding the added SNOMED concepts